\pdfoutput=1

\documentclass[11pt]{article}

\usepackage[final]{acl}

\usepackage{times}
\usepackage{latexsym}
\usepackage[T1]{fontenc}
\usepackage[utf8]{inputenc}

\usepackage{microtype}

\usepackage{inconsolata}

\usepackage{graphicx}

\usepackage{booktabs} %
\usepackage{multirow}

\title{Gla-AI4BioMed at RRG24: Visual Instruction-tuned Adaptation for Radiology Report Generation}

\author{Xi Zhang, Zaiqiao Meng\Thanks{~Corresponding author.}, Jake Lever \and Edmond S. L. Ho\\
  School of Computing Science, University of Glasgow \\
  \texttt{X.Zhang.6@research.gla.ac.uk, Zaiqiao.Meng@glasgow.ac.uk}\\
  \texttt{Jake.Lever@glasgow.ac.uk, Shu-Lim.Ho@glasgow.ac.uk} \\}

\begin{document}
\maketitle
\begin{abstract}
We introduce a radiology-focused visual language model designed to generate radiology reports from chest X-rays. Building on previous findings that large language models (LLMs) can acquire multimodal capabilities when aligned with pretrained vision encoders, we demonstrate similar potential with chest X-ray images. This integration enhances the ability of model to understand and describe chest X-ray images. Our model combines an image encoder with a fine-tuned LLM based on the Vicuna-7B architecture, enabling it to generate different sections of a radiology report with notable accuracy. The training process involves a two-stage approach: (i) initial alignment of chest X-ray features with the LLM (ii) followed by fine-tuning for radiology report generation\footnote{\url{https://github.com/Glasgow-AI4BioMed/RRG-BioNLP-ACL2024}.}. 

\end{abstract}

\section{Introduction}
Radiology reports constitute the primary medium through which radiologists convey the findings and conclusions derived from radiography, such as chest X-rays. These reports play a pivotal role in the diagnostic and therapeutic processes across a wide range of diseases, emphasizing their significance in contemporary medical practice \citep{engle2021evidence}. Structured to enhance clarity and efficacy in medical communication, radiology reports primarily feature FINDINGS and IMPRESSIONS sections \citep{aaf10e715ced41219bde7e011e1e602c}. The FINDINGS section details the critical observations of the radiologist on the image, while the IMPRESSIONS section summarizes the conclusions and recommendations of the radiologist. These sections collectively ensure that radiology reports are indispensable in diagnostic and therapeutic decision-making, combining image analysis and clinical insight. Table~\ref{table1} shows an example generated by GPT-4~\cite{openai2024gpt4}, which delineates these sections.

\begin{table}[ht]
  \centering
  \normalsize
  \begin{tabular}{p{7cm}}
    \hline
    \hline
    \textit{FINDINGS} \\
    \hline
    There has been an increase in size of the left pleural effusion compared to the prior exam. The right lung remains clear with no evidence of consolidation or pneumothorax. The heart size is mildly enlarged but stable. The mediastinum appears unremarkable. Mild degenerative changes are noted in the thoracic spine and ribs. The upper abdomen is without remarkable findings. \\
    \hline
    \hline
    \textit{IMPRESSIONS} \\
    \hline
    Increase in left pleural effusion compared to prior. Stable mild cardiomegaly. No evidence of right lung pathology. \\
    \hline
  \end{tabular}
  \caption{FINDINGS and IMPRESSIONS in a synthetic radiology report generated by GPT-4.}
  \label{table1}
\end{table}

Radiology report generation (RRG) is crucial for advancing future medical artificial intelligence systems \citep{monshi2020deep}. This task involves transforming images into text, necessitating alignment between imaging and textual data. Significant advancements in natural language processing have driven progress in this area, with large generative visual language models like LLaVA \citep{liu2023visual}, InstructBLIP \citep{dai2023instructblip}, and Flamingo \cite{alayrac2022flamingo} leading the way.

The prevailing visual language models, such as those mentioned above, aim to address the challenge of multimodal alignment by leveraging large-scale pretraining. Typically, this involves adapting a vision encoder for integration with a pretrained LLM. To meet specific task requirements, various degrees of finetuning are applied. For example, LLaVA \citep{liu2023visual} represents a novel end-to-end trained large multimodal model for general-purpose visual and language understanding, achieving impressive chat capabilities. However, in the context of our work, the focus is on fine-tuning image-text pairs, specifically for tasks related to medical images, to enhance the capability of the visual language model in radiology report generation.

In the specialized area of radiographic report generation, it is paramount for models to discern nuanced details within multiple medical images. These details include subtle variations in opacity against a backdrop of overlapping structures \citep{9103969}. Therefore, the radiographic report generation task extends beyond the mere extraction of details from a single image. Models must interpret the clinical implications of these nuances to generate precise and medically rigorous text in reports. This is a particularly crucial capability for radiographic report generation models, since it enhances the clinical utility and effectiveness of radiology reports and ensures their accuracy and relevance in clinical settings.

General-domain models have proven inadequate for generating findings in radiology reports \citep{hyland2024maira1}. In this work, we propose a radiology-specific visual language model designed for solving the radiology report generation task by fine-tuning across various sections of medical reports. Our model utilizes a two-stage fine-tuning process that significantly enhances its performance. In particular, we initially align the large language model with image embedding through a pretraining phase. In the second stage, we further fine-tune the LLM using Low-Rank adaptation (LoRA) techniques \citep{hu2021lora}. Both stages are trained on the dataset in this workshop \citep{xu-etal-2024-overview}. 

Additionally, we use a straightforward strategy of merging and stitching multiple images to form a single cohesive input, enabling the model to effectively process and integrate information from multiple X-ray images. Using the dataset provided by this workshop, which includes a collection of chest X-rays and their corresponding sections, we fine-tune our model to enhance the accuracy and specificity of the generated radiology reports.

This paper investigates the fine-tuning of visual instruction for a visual language model in the specific domain of radiology report generation. We describe the training of two distinct models developed for the Shared Task on Large-Scale Radiology Report Generation (RRG24) at the BioNLP 2024 Workshop \citep{xu-etal-2024-overview}. In the public test set, we achieved an F1-RadGraph score \citep{delbrouck-etal-2022-improving} of 24.13 and 22.79 in the Findings and Impressions sections, respectively. In the hidden test set, we achieved F1-RadGraph scores \citep{delbrouck-etal-2022-improving} of 24.13 and 22.10 in the Findings and Impressions sections, respectively, which places us 4th on the leaderboard at the time of submission. The contributions of this research are as follows:

\begin{itemize}
  \item We enhance domain adaptation for radiology by implementing visual instruction tuning, which further fine-tunes the visual language model specifically for image-to-text tasks. This approach optimizes performance in interpreting and translating visual data into descriptive, clinically relevant text.
  \item We adopt a method of stitching multiple images together, allowing a single image encoder to process multiple image inputs simultaneously. This strategy obviates the need for separate encoding of each image, enabling the model to adapt to varying numbers of image inputs using limited resources.
\end{itemize}

\section{Related Work}

Nowadays, exemplified by open-source projects such as LLaVA \citep{liu2023visual}, the effectiveness of self-supervised vision-language models (VLMs) using parallel data has been demonstrated in different research domains. These VLMs, when instruction-tuned with multimodal inputs, align well with human intentions and perform robustly in various downstream tasks, including converting images to text \citep{park2023visual}.

However, the unique characteristics of biomedical image-text pairs significantly differ from those in general domains. Biomedical images often contain subtle and complex features that require precise interpretation, while the corresponding text must convey highly specific medical information \citep{huff2021interpretation}. In biomedical settings, VLMs designed for general domains often fail to meet these specialized needs, as they lack the ability to accurately interpret medical data and generate relevant clinical descriptions \citep{chang2023survey}. This discrepancy underscores the urgent need for domain-specific fine-tuning. By tailoring VLMs to the distinct demands of the biomedical field, such fine-tuning can enhance their ability to capture and convey the intricate details necessary for accurate medical interpretations and reports.

Recent advancements have been made in adapting general-purpose foundation models for medical applications, particularly in radiology. The Med-Flamingo \citep{moor2023medflamingo}, an extension of the OpenFlamingo framework \citep{awadalla2023openflamingo}, leverages images and captions from medical textbooks to enhance few-shot visual question-answering capabilities. Similarly, the Med-PaLM M, developed by \citet{tu2023generalist}, fine-tuned the PaLM-E model \citep{driess2023palme} using comprehensive biomedical datasets. LLaVA-Med, proposed by \citet{li2023llavamed}, modifies the LLaVA \citep{liu2023visual} framework with image-text pairings and multimodal instructions from PubMed data. Additionally, the ELIXR model, developed by \citet{xu2023elixr}, integrates the SupCon CXR encoder \citep{doi:10.1148/radiol.212482} with the PaLM 2-S model \citep{anil2023palm} to support classification, semantic search, question answering, and quality assurance. Finally, the Radiology-GPT, created by \citet{liu2024radiologygpt}, utilizes radiology reports from MIMIC-CXR \citep{johnson2019mimic} to facilitate the generation of findings-to-impression text, based on the Alpaca instruction-tuning framework \citep{taori2023alpaca}.

Historically, research in radiology report generation has varied, with some studies focusing exclusively on either the Findings or the Impressions sections \citep{jin2024promptmrg,yan2023styleaware}, while others have addressed both. Notably, \citet{endo2021convolutional} and \citet{bannur2023learning} specialized in generating only the Impressions section. In contrast, studies by \citet{miura-etal-2021-improving}, \citet{delbrouck-etal-2022-improving}, \citet{Tanida_2023}, \citet{Nicolson_2023}, and \citet{tu2023generalist} concentrated on the Findings section. Comprehensive analyses by \citet{yu2023evaluating} and \citet{jeong2023multimodal} covered all settings, demonstrating that the choice of sections significantly influences reported performance metrics, complicating comparative evaluations across different study designs.

However, these existing models have limitations. Most notably, they are typically designed to process single images and often fall short in generating reports that match the depth and detail of those written by human radiologists. Additionally, they do not fully replicate the workflow of medical professionals, who often reference multiple images to enhance report accuracy. Our work addresses these gaps by developing a model capable of handling multiple images simultaneously and generating comprehensive radiology reports. This approach aims to more closely mimic the process used by medical professionals, thereby improving the accuracy and quality of the generated reports.

\section{Methodology} 
In our study, we follow the observations from LLaVA-Med \cite{li2023llavamed}, suggesting superior performance when initiating with a language-only pretrained LLM rather than a multimodal-trained base. Our model architecture incorporates an image encoder and a learnable adapter placed atop the image outputs, mirroring the LLaVA-1.5 model design \citep{liu2023visual}. We adopt an auto-regressive language modelling approach using cross-entropy loss \citep{graves2014generating,long2024clceapproachrefiningcrossentropy} and align hyperparameters with those from LLaVA-1.5, including a joint tuning phase for the LLM and adapter \citep{liu2023visual}. In alignment with LLaVA-1.5 protocols, we initially pretrain the adapter alone for one epoch, followed by a full training cycle lasting three epochs, employing Low-Rank Adaptation of Large Language Models techniques (LoRA) \citep{hu2021lora} for efficient parameter tuning.

\begin{figure*}[ht]
  \includegraphics[width=\linewidth]{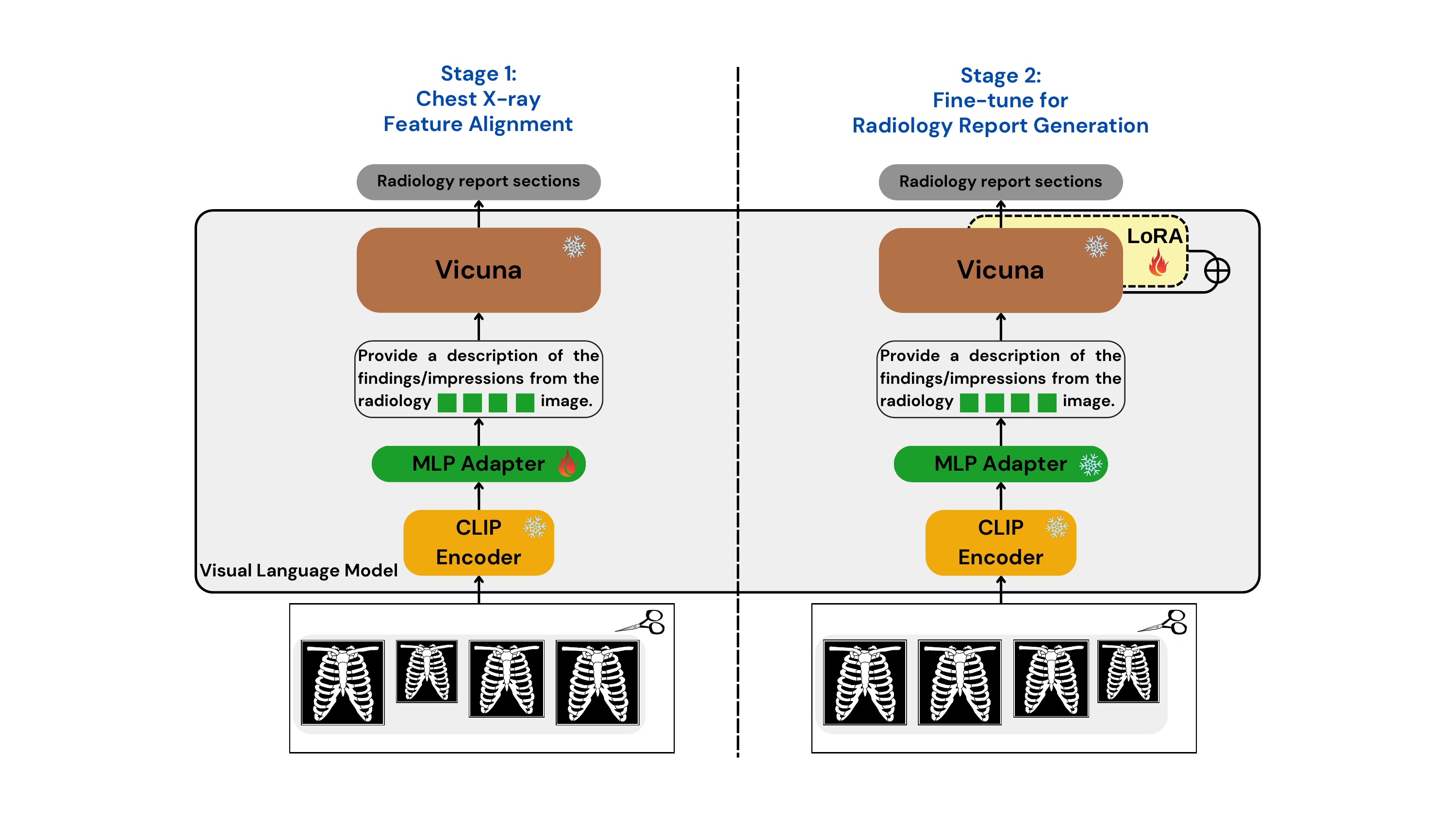}
  \caption{Our two-stage training framework. In the first stage, visual features are aligned with LLM. In the second stage, the model focuses on the training of radiology report generation tasks.}
  \label{fig:framework}
\end{figure*}

\subsection{Task Description}

A key application of natural language generation in medicine is developing support systems that produce written reports from X-ray images, detailing clinical findings. Such systems are highly valuable, potentially reducing the routine workload of radiologists and improving the efficacy of clinical interactions. The objective of this shared task is to generate radiology reports from one or more chest X-rays taken during a single study, specifically targeting two sections: `Findings' and `Impressions' (as shown in Table~\ref{table1}). 

Consequently, our team is dedicated to the task of exclusively producing either the `Findings' or `Impressions' sections of the report. We have developed separate models for each section because their focuses are different. The `Findings' section provides a factual description based on the images, while the `Impressions' section offers the radiologist's conclusions and recommendations. By separating the models, we can tailor each to better address its specific requirements.

For the radiology report section generation, handling multiple images is crucial as it allows the model to provide a detailed and accurate description of the observed facts, similar to how radiologists analyze multiple images to form a comprehensive understanding. This approach enhances the model's ability of to mimic the actual workflow of medical professionals, who often reference multiple images.

\subsection{The Proposed Models}
To solve the RRG24 shared task, we fine-tune a large visual language model for radiology report generation based on the provided dataset. Specifically, we propose two separate models, called Med-CXRGen-F and Med-CXRGen-I, fine-tuned on Findings and Impressions sections respectively.

We use CLIP \citep{radford2021learning} as an image encoder and Vicuna-1.5 \citep{vicuna2023} as a large language model. Our adaptation module consists of a multi-layer perceptron (MLP) featuring GELU activations \citep{hendrycks2023gaussian} and a uniform hidden size of 1024 across all layers. 

The interaction with the model involves alternating system messages linked with the corresponding image. The training objective of the model is to generate accurate responses. Initially, we convert the image into a series of image patch tokens via the image encoder, selecting embeddings from the penultimate layer. These image features are then processed by the MLP adapter, aligning them to the input specifications of the LLM. 

The instructional prompt of the report generation task we employed is: "Provide a description of the findings/impressions from the radiology <image>\verb|\n| image." In this prompt, "<image>\verb|\n|" represents the image holder token, as shown in Figure~\ref{fig:framework}, which indicates to the LLM that it should base its generation on the input image.

\subsection{Training}\label{sec:meth}
The same network architecture is utilized for different radiology report sections, where an MLP adapter connects the vision encoder and the language model. For model training, we use a two-stage procedure: (as shown in Figure~\ref{fig:framework})

\begin{itemize}
\item \textbf{Stage 1: Chest X-ray Feature Alignment} \\ In the first epoch training phase on the provided dataset, each sample, accompanied by instructions and image input, prompts the model to predict the original caption. During this stage, we keep the visual encoder and LLM weights unchanged, focusing solely on updating the MLP adapter. This approach aligns the features from chest X-ray images with their textual embeddings in the LLM. Training is limited to a single epoch, which facilitates the expansion of the vocabulary of aligned image-text tokens specific to the radiology domain.
\clearpage
\item \textbf{Stage 2: Fine-tune for Radiology Report Generation} \\ In the second phase, the visual encoder weights and adapter are kept frozen while continuing to update the pre-trained LLM weights using LoRA \citep{hu2021lora} technology. Further fine-tuning is conducted on the provided dataset through visual instrumental tuning with three epochs.
\end{itemize}

\section{Evaluation}
\subsection{Dataset}

We fine-tune and evaluate our models using the RRG24 dataset hosted on the BioNLP ACL'24 \citep{xu-etal-2024-overview}, which includes data from MIMIC-CXR \citep{johnson2019mimic}, CheXpert \citep{chambon2024chexpert}, PadChest \citep{Bustos_2020}, BIMCV-COVID19 \citep{vaya2020bimcv}, and OpenI, with their statistics shown in Table~\ref{table2}.

\begin{table}[ht]
  \centering
    \normalsize
    \begin{tabular}{lll} 
    \toprule
    Dataset    & FINDINGS          & IMPRESSIONS       \\
    \midrule
    training   & 344,394   & 366,413     \\
    validation  & 8,839     & 9,331     \\
    test-public  & 2,692     & 2,967     \\
    test-hidden  & 1,063	     & 1,428     \\
    \bottomrule
    \end{tabular}
  \caption
    {Distribution of shared task on Large-Scale Radiology Report Generation.}
  \label{table2}
\end{table}

We conducted training in two stages (refer to section \ref{sec:meth}). To ensure consistency between training and inference processes, we analysed the word count distribution, as shown in Table~\ref{table3}. Consequently, we have set a maximum length of 1024 for both the text input and inference output to minimise computational expense. On the other hand, as illustrated in Table~\ref{table4}, some datasets contain multiple images, therefore, we select up to the first four images for the image input. We merge multiple images horizontally to form a single-image input, which is proven to be robust in our experiments.
\begin{table}[ht]
  \centering
  \normalsize
    \begin{tabular}{lcc} 
    \toprule
    Dataset    & FINDINGS          & IMPRESSIONS       \\
    \midrule
    training   & 259 (±180)   & 216 (±153)     \\
    validation  & 257 (±176)     & 217 (±155)    \\
    test-public  & 380 (±161)     & 257 (±224)    \\
    \bottomrule
    \end{tabular}
  \caption{Average word count and standard deviation on Large-Scale Radiology Report Generation.}
  \label{table3}
\end{table}

\begin{table}[ht]
  \centering
  \normalsize
    \begin{tabular}{lcc} 
    \toprule
    Dataset    & FINDINGS          & IMPRESSIONS       \\
    \midrule
    training   & 1.57 (±0.63)   & 1.45 (±0.62)     \\
    validation  & 1.58 (±0.62)     & 1.45 (±0.62)    \\
    test-public  & 1.70 (±0.71)     & 1.67 (±0.71)    \\
    \bottomrule
    \end{tabular}
  \caption{Average number of images and standard deviation on Large-Scale Radiology Report Generation.}
  \label{table4}
\end{table}

\subsection{Metrics}
We assess the generated reports through a dual approach involving both general lexical metrics and specialized radiology metrics. Focusing on the accuracy of described medical findings, radiology-specific metrics provide a deeper insight into the clinical relevance of the reports, beyond surface-level phrasing variations. According to the RRG24 guidelines, we consider five evaluation metrics for this work, including BLEU4 \citep{10.3115/1073083.1073135}, ROUGEL \citep{lin-2004-rouge}, BERT score \citep{zhang2020bertscore}, F1-cheXbert \citep{smit2020chexbert}, and F1-RadGraph \citep{delbrouck-etal-2022-improving}.

\begin{table*}[ht]
  \centering
  \small
  \begin{tabular}{llcccccc}
    \toprule
    Model & Dataset & Section & BLEU4 & ROUGEL & Bertscore & F1-cheXbert & F1-RadGraph \\
    \midrule
    \multirow{3}{*}{Med-CXRGen-F} 
    & validation & Findings & 7.02 & 23.33 & 48.93 & 40.42 & 21.94 \\
    & test-public & Findings & 8.07 & 24.90 & 53.45 & 45.91 & 24.13 \\
    & test-hidden & Findings & 7.65 & 24.35 & 52.69 & 46.21 & 24.13 \\
    \midrule
    \multirow{3}{*}{Med-CXRGen-I} 
    & validation & Impressions & 10.18 & 28.10 & 51.78 & 50.51 & 26.65 \\
    & test-public & Impressions & 7.10 & 25.11 & 47.39 & 47.43 & 22.79 \\
    & test-hidden & Impressions & 9.60 & 25.27 & 48.60 & 46.74 & 22.10 \\
    \bottomrule
  \end{tabular}
  \caption{Evaluation results on different datasets.}
  \label{table5}
\end{table*}

\subsection{Training details}
We evaluated our two proposed models, i.e. Med-CXRGen-F and Med-CXRGen-I, on the workshop evaluation datasets, based on a computational infrastructure utilizing an A6000 GPU (48GB memory each) with the Deepspeed zero-3 configuration \citep{rajbhandari2020zero} with BF16 enabled. We employ a cosine learning rate scheduler that begins with a warm-up phase of 0.03 and sets the learning rate at $1 \cdot 10^{-5}$. The global batch size for our experiments is set at 16. Observations of the smallest loss on the evaluation dataset throughout the training process guide us to select this as the final checkpoint for all runs. For inference on the test dataset, we decode in 32-bit precision up to 150 tokens, consistent with the baseline model on the leaderboard \citep{delbrouck-etal-2022-vilmedic}. Each model required approximately 215 hours of training. 

\section{Results}
We report the performance of our two proposed models over five evaluation metrics in Table ~\ref{table5}. As shown in Table~\ref{table5}, in the public test set, we achieved an F1-RadGraph score \citep{delbrouck-etal-2022-improving} of 24.13 and 22.79 in the Findings and Impressions sections, respectively. In the hidden test set, we achieved F1-RadGraph scores \citep{delbrouck-etal-2022-improving} of 24.13 and 22.10 in the Findings and Impressions sections, respectively, which places us 4th on the leaderboard\footnote{https://vilmedic.app/misc/bionlp24/leaderboard} at the time of submission. Additionally, our model achieved notable Bertscore results in the test-public set, with 53.45 for Findings and 47.39 for Impressions. These results demonstrate the effectiveness of our approach in generating high-quality medical reports across different datasets.

\section{Discussion}
Performance disparities observed between the Findings and Impressions sections of the test results can be attributed to several factors. Firstly, the Impression and Findings sections address distinct medical purposes, resulting in performance disparities. The Findings section offers an objective description of symptoms, while the Impressions are oriented towards diagnostic conclusions. The variability in word count between these sections also affects the complexity of model inference, as reflected in the lexical evaluation scores.

Additionally, significant discrepancies in the medical evaluation metrics highlight a varied distribution of diseases within the test set. This heterogeneity could impact the generalisability and accuracy of the model. Furthermore, our analysis indicates that the performance may be compromised in multi-image inference scenarios where it does not account for superfluous images. Such factors are essential to consider when assessing the diagnostic accuracy and reliability of the model in clinical settings. Enhancing the ability of model to differentiate between relevant and superfluous images could significantly improve diagnostic accuracy. 

Furthermore, exploring domain-specific adaptations and fine-tuning strategies tailored to the unique characteristics of medical data could further enhance model performance. Incorporating temporal dynamics into the model to capture changes over time and developing more sophisticated frameworks for generating multi-modal radiology reports are other promising avenues for future research. These advancements are expected to enhance both the practicality and accuracy of our model within clinical scenarios.

\section{Conclusion}
In this work, we have developed a vision-language model capable of processing multiple images. Through visual instruction tuning, we achieved alignment between two modalities and further fine-tuning for specific downstream tasks. Notably, our system attained a commendable fourth-place standing across four diverse test datasets at the RRG24 at BioNLP 2024 workshop \citep{xu-etal-2024-overview}, substantiating the practicality of vision-language models within specialized medical tasks.

Moving forward, we intend to conduct in-depth research into more sophisticated methods for generating multi-modal radiology reports. This will involve incorporating temporal dynamics and developing frameworks specifically focused on text generation. Such advancements are expected to enhance both the practicality and accuracy of our model within the clinical scenario.
\clearpage
\section{Limitations}
The following section outlines the limitations identified in our study:

\begin{enumerate}
\item \textbf{Prevalence of certain conditions:} Some diseases are more easily detected, which may lead to artificially high medical assessment scores.

\item \textbf{Imaging modalities and anatomical structures:} There is a notable imbalance in the imaging modalities and anatomical structures covered in the training dataset. Variations such as the number of images per patient and the considerable disparity in the length of medical reports exacerbate this imbalance.

\item \textbf{Radiologist and radiology department preferences:} Preferences and writing styles vary among radiologists and radiology departments. This diversity adds complexity to medical reports by introducing inconsistencies and uncertainties that are, to a certain extent, human-induced. For example, the dataset provided in this workshop demonstrates that even the same radiology section descriptions have varying styles. These elements significantly complicate the task of report generation.
\end{enumerate}
These limitations highlight areas for improvement and the need for methodological refinements to enhance model effectiveness and reliability in clinical environments. These challenges were not addressed within the scope of this workshop.

\section*{Ethics Statement}
The concepts and methodologies presented in this paper are based on experimental research findings. They are not currently available for other use. The data used for the training processes were exclusively sourced from the provided dataset, which complies with ethical standards regarding Patient Health Information. Adherence to these standards guarantees the responsible handling of sensitive data throughout our research.

\section*{Acknowledgments}
We extend our gratitude to the BioNLP 2024 RRG24 Shared Task organizers for providing the baseline pipeline ViLMedic \citep{delbrouck-etal-2022-vilmedic} and curating these challenging and exciting tasks \citep{xu-etal-2024-overview}.

\bibliography{custom}

\end{document}